# Diagnóstico de la estrategia editorial de medios informativos chilenos en Twitter mediante un clasificador de noticias automatizado

## Diagnosing editorial strategies of Chilean media on Twitter using an automatic news classifier

**Resumen**

En Chile no existe una instancia independiente que publique estudios cuantitativos o cualitativos constantes para comprender el ecosistema de medios tradicionales y sus adaptaciones en la Web Social. Los públicos consumen informaciones ya no sólo en periódicos y noticieros, sino también a partir de redes sociales como su fuente primaria de información. Twitter es la red social de noticias por excelencia y los medios hacen esfuerzos por ganar adeptos en esta red social. Es este artículo se propone una metodología basada en minería de datos web. Utilizamos técnicas de rastreo y extracción de flujos de noticias de 37 medios de comunicación chilenos que presentan una vida activa en Twitter y proponemos varios indicadores para compararlos. Analizamos los volúmenes de producción, sus audiencias potenciales y , usando técnicas de procesamiento natural del lenguaje, exploramos el contenido de la producción informativa: sus tendencias editoriales y cobertura geográfica.

**Abstract**

In Chile, does not exist an independent entity that publishes quantitative or qualitative surveys to understand the traditional media environment and its adaptation on the Social Web. Nowadays, Chilean newsreaders are increasingly using social web platforms as their primary source of information, among which Twitter plays a central role. Historical media and pure players are developing different strategies to increase their audience and influence on this platform. In this article, we propose a methodology based on data mining techniques to provide a first level of analysis of the new Chilean media environment. We use a crawling technique to mine news streams of 37 different Chilean media actively presents on Twitter and propose several indicators to compare them. We analyze their volumes of production, their potential audience, and using NLP techniques, we explore the content of their production: their editorial line and their geographic coverage.

## 1 Introducción

En el ecosistema de medios informativos chileno, interactúan más de 2000 concesiones de radio (Ramírez 2009), más de 100 canales de televisión con propuestas comunicacionales y cerca de 90 periódicos de prensa escrita tradicional (Corrales y Sandoval 2005). Estas cifras pueden tomarse como un punto de partida, pero si consideramos la existencia de canales locales de cable y la emergencia de medios online nos encontramos ante un escenario en expansión y poco sistematizado. En este estado del arte resultan interesante de consignar investigaciones como el *Primer Estudio Nacional sobre Lectoría de Medios Escritos* de Azocar et al. (2010) y *Diarios de vida de las audiencias chilenas* de Arriagada et al. (2014).

Sabemos que en el nuevo ecosistema de medios conviven medios tradicionales con sus señales de televisión o radio y periódicos impresos, pero además están los medios online entre los que cohabitan las versiones digitales de los medios tradicionales y nuevos medios nacidos para la red de redes con posterioridad al año 2000[1]. Los nuevos medios, además de poner sus contenidos en formatos web, han debido adaptarse constantemente a las evoluciones tecnológicas que se proponen desde nuevas plataformas de redes sociales, nuevos dispositivos móviles y nuevos recursos de *broadcasting* digital.

Además de la poca sistematización de los cambios en el espacio digital chileno, existe un conocimiento muy limitado del alcance de los medios de comunicación y la generación de audiencias. En países como España o Colombia existen organismos destinados ha realizar seguimientos independientes de la realidad medial (http://www.aimc.es y http://www.acimcolombia.com). En cambio, en Chile, si bien en forma permanente se están haciendo mediciones de *people meter* entre los grandes canales de televisión, dichas evaluaciones se hacen a través de dispositivos distribuidos en siete ciudades (http://www.ibopemedia.cl) y poco pueden decir de los canales de televisión locales. En la radiofonía, las evaluaciones también son parciales ya que, aunque la cobertura de regiones es menos centralizada y da lugar a las identidades locales, las estadísticas existentes provienen de estudios desarrollados por IPSOS[2] por encargo por ARCHI (Asociación de Radiodifusores de Chile) que agrupa aproximadamente a un 50% de un total de más de 2000 concesiones. Por su parte, los estudios de circulación y lectoría de periódicos evalúan sólo la realidad del gran Santiago y en contadas ocasiones miden realidades de grandes ciudades como Valparaíso y Concepción, pero poco se sabe del impacto en el resto de las ciudades del país. Además, los estudios son encargados por la ANP (Asociación Nacional de la Prensa) que reúne a los medios más tradicionales y poderosos de Chile y tienen por finalidad aportar información a los agentes publicitarios[3].

Cabe señalar además que en Chile existe una particular concentración de la propiedad de los medios que trae consigo una expresión limitada del pluralismo editorial y la escasa expresión de las realidades locales. Por otra parte, algunos estudios esporádicos de universidades u organismos independientes han dado cuenta de: a) un nivel fuerte de centralismo en las informaciones (Stambuk 1999; Puente y Grassau 2009), y de b) falta de pluralismo y expresión social en los medios (Vera 2005; Vergara et al. 2012). Así por ejemplo el Consejo Nacional de Televisión (CNTV, 2014) concluye que el 49% de los encuestados entre 18-24 años de edad consideraron que las informaciones resultan insuficientes para entender las movilizaciones estudiantiles. Es en esos contextos, cuando los medios no responden a las demandas comunicativas de la sociedad, que las opiniones en redes sociales emergen con potencia. En alguna medida se desarrollan procesos comunicativos sobre las opiniones de multitudes en un sentido similar a lo que Baeza-Yates y Sáez-Trumper (2015) definen como sabiduría de multitudes en redes sociales.

[1] En el año 2000 nace "El mostrador" como periódico electrónico nacido solo en formato electrónico. En el año 1994, nace la tercera online como espejo de su versión tradicional en papel.

[2] Ipsos en Chile desarrolla tradicionalmente los informes de radiofonía: http://www.ipsos.cl/ipsosradioalaire/pagcuatro.htm

[3] Asociación Chilena de Agencias de Publicidad: http://www.achap.cl/estudios.php

## 2. Justificación y estado del arte

### 2. 1. Cambios en el contexto digital

En un contexto en el que por una parte la sociedad chilena presenta una fuerte adherencia al uso de redes sociales y los medios de comunicación configuran estrategias específicas de *community manager* para redes sociales, resulta interesante desarrollar una herramienta que permita escanear el ecosistema informativo chileno, categorizar la información circulante y visualizar características del flujo informativo chileno. Por ello, desarrollamos una serie de herramientas que permiten observar el flujo de informaciones de 37 medios informativos[4] desde la red social Twitter. Entre los medios seguidos se encuentran *"medios tradicionales"* como El Mercurio, La Tercera, Mega, TVN, entre otros que nacieron previos a internet y se adaptaron a la misma y "*nuevos medios*" tales como El mostrador, El Desconcierto, Pulso que nacieron exclusivamente en internet.

Imaginemos que la forma primera de informarse son las redes sociales. Cada mañana los ciudadanos revisan la prensa desde Twitter o Facebook y valoran, comentan y comparten informaciones hacia sus círculos sociales. Desde ahí, y a lo largo del día, se generan lecturas y opiniones que se encuentran mediadas ya no sólo por los medios de prensa que aportaron las informaciones, sino también por las interacciones viralizadoras que familiares, amigos o conocidos ejecutan seleccionando, reponiendo y comentando noticias en las redes sociales (RRSS). Con todo, internet ha cambiado la forma de informarse de los ciudadanos y el poder de la audiencias y los grandes medios se están reconfigurando con nuevas reglas (Sáez-Trumper 2011).

Si bien no se puede asumir que ese sea el mecanismo mayoritario de informarse para toda la población, si existen antecedentes concretos (Newman 2013) de que en el mundo se construye una tendencia en la que los sujetos menores de 35 años consideran que internet es la mejor forma de acceder a noticias y el grupo sobre esa barrera etaria (mayores de 35 años de edad) tiende preferir informarse más tradicionalmente vía televisión. Además se puede inferir que probablemente, con el paso de las generaciones, la frontera de 35 años se ira desplazando.

Chile a demostrado ser un país que adopta con facilidad las tendencias derivadas de las Tecnologías de la Información y las Comunicaciones (TICS) y tanto ComScore (2013) como IAB (2012) señalan que la penetración del *social media* alcanza aproximadamente al 80% de la población. Estudios recientes de Carcamo-Ulloa y Saez-Trumper (2014) dan cuenta de que las interacciones con medios de prensa en Facebook pueden tener un volumen similar en España que en Chile, con la salvedad de que la población de nuestro país representa aproximadamente un 35% de la de España. Además, Halpern (2014) detecto en Chile una interesante tendencia al consumo de informaciones interactuando entre pantallas (por ejemplo, ver TV y utilizar RRSS al mismo tiempo).

[4] Esta cifra de medios responde a la etapa inicial del proyecto, pero en la actualidad ya se pueden seguir más de 250 medios de comunicación chilenos.

Los nuevos ecosistemas de la información (Diaz-Nosty 2013) incluyen como agentes relevantes de comunicación a las redes sociales. Cardoso (2014) señala que los "medios sociales" han cambiado el modelo de comunicación, "no solo porque técnicamente han multiplicado las formas en que podemos apropiarnos de la comunicación en cuestión o elegir ~~a~~ relacionarnos de forma determinada, sino también porque son el eslabón perdido necesario para operar una revolución en nuestro modelo de comunicación" (Cardoso 2014: 18). Por su parte, los medios de prensa tradicionales también incorporan herramientas como Facebook y Twitter a modo de soportes divulgativos de las informaciones que cuelgan en los *website* (Tejedor-Calvo 2010). Todo ello con el fin de viralizar sus contenidos informativos o de entretenimiento. Además, en el día a día las rutinas de los medios de prensa hacia sus redes sociales resultan más o menos reproductivas de sus informaciones y aparece en las redacciones la figura del *Community Manager* o el *Social Media Editor* (Sánchez & Méndez, 2013). Sin embargo, gran parte de la actividad viral de las informaciones depende también de las interacciones que los propios lectores/usuarios ejecutan valorando *(like)*, comentando *(comment)* o compartiendo *(share)* contenidos desde y hacia redes sociales como Facebook (Cabalin 2014) o del retweet en Twitter.

Esta última situación trae consigo un cambio importante a nivel de audiencias, pues, por un lado, los liderazgos informativos en redes sociales pueden no reproducir exactamente los ranking de lectorías y/o audiencias tradicionales (Carcamo-Ulloa y Saez-Trumper 2013), mientras que, por otro lado, permitirían la irrupción nuevos medios alternativos y c) un medio en internet deja de ser sólo una radio, un periódico o una televisora y se convierte en un difusor de distintos contenidos informativos multimodales (Pardo 2012a, 2012b) o hipertextuales, interactivos y multimediales (Masip et al. 2010). Una prueba de ello es que hoy el contenido más viralizado de una radio emisora online puede ser un video recuperado de YouTube o aportado por un reportero ciudadano.

## 2.2. ¿Por qué la minería de datos web?

La herramienta para describir un mapa real de los medios en el nuevo ecosistema informativo de las redes sociales está, y tiene más de tres décadas de desarrollo de la mano de los investigadores en ciencias de la computación: se trata de la minería de datos web. Dichos estudios están instalados en áreas como la bibliotecología (Peters 1993; Kurth 1993; García, Botella y Marcos 2010; Marcos y González 2010) y también en los entornos virtuales adaptativos (Morales, Soto y Martínez 2005) existe una valoración creciente del *Web Mining* como aporte a la búsqueda de patrones para el mejoramiento de los procesos comunicacionales involucrados. En tal sentido, Baeza-Yates (2009) y Baeza-Yates y Passi (2011) hacen una panorámica que revisa las principales tendencias de expansión para la minería de datos en la Web 2.0, el spam, análisis de búsquedas, redes sociales y la privacidad en internet.

Para el ámbito de la comunicación social existen experiencias relevantes en el análisis apoyado en la minería de datos en Europa y Estados Unidos (Yang y Leskovec 2011) y que aportan una perspectiva internacional (Sáez-Trumper, Castillo y Lalmas 2013). Sin embargo, hacen falta diagnósticos locales que en Chile aún no se desarrollan dado que la expansión de los medios de comunicación en los últimos años es creciente y no se conocen sus niveles de influencia en redes sociales. Así por ejemplo, en el catastro realizado por *mediaonline.net* se describen 296 medios para el caso chileno. Ellos se distribuyen irregularmente en la geografía y con una lógica de mayor concentración en la Región Metropolitana y sus regiones vecinas, y menor despliegue hacia los extremos norte y sur del país. Conociéndose muy poco sobre los impactos de la producción de información de los medios alternativos y/o regionales y sus potenciales audiencias.

Pese a que, evidentemente, Chile no presenta una realidad de cultura digital homogénea y que existen comunidades que aun se encuentran aisladas de las sociedad de la información, esta claro que la brecha digital esta hoy en los usos más que en los accesos. Es por eso que el presente trabajo tiene por objetivo proponer un modelo de análisis basado en minería de datos en redes sociales como diagnóstico de los comportamientos de los usuarios de medios de prensa en la internet chilena y proyectar usos educativos de esa información. Esta propuesta entremezcla estrategias metodológicas basadas en las minería de datos web (Baeza-Yates 2009), específicamente seguir las interacciones en el *social media* (Carcamo-Ulloa y Saez-Trumper 2013), hacer minería de textos para clasificar estas noticias (Saez-Trumper, Castillo y Lalmas 2013; Vernier, Monceaux y Dille 2012), y la detección de temas emergentes en Twitter (Guzmán y Poblete 2013).

## 3. Metodología del estudio

### 3.1. Datos y procedimientos de análisis

Se realizó un procedimiento que permitió seguir todas las informaciones emitidas en Twitter por 37 medios de comunicación chilenos. El *crawler* inicial almacena las informaciones en una base de datos no SQL y como cada tweet contiene un hipervínculo a la información extendida en una página web también se realiza un segundo proceso de *crawling to link* a la noticia para aplicar una limpieza o *scraping* de contenido relevante de la web. Para construir la infraestructura de minería y análisis de datos se recurrió a herramientas tales como: Elasticsearch, Kibana y UIMA.

*Figura 1.* Representación del modelo de obtención de datos

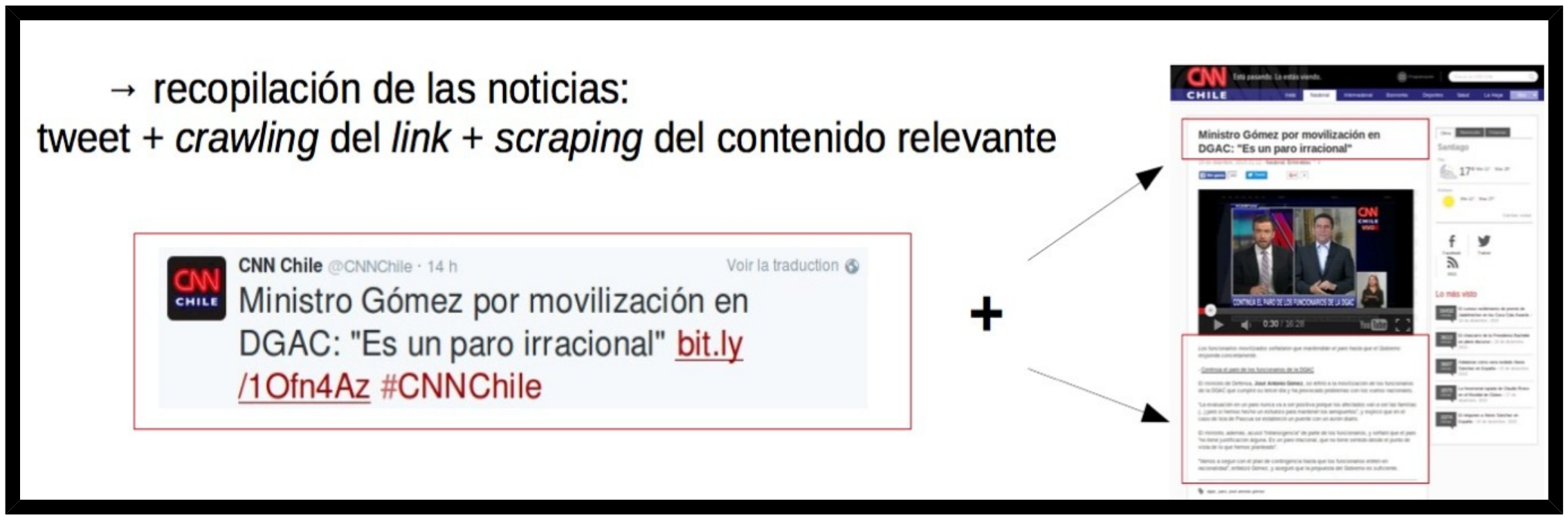

Los datos se obtuvieron entre el 01 de junio de 2015 y el 30 de noviembre de 2015 para contemplar un semestre móvil. Los medios emitieron vía sus cuentas de Twitter 736.538 noticias: junio, 120571; julio, 119704; agosto, 124910; septiembre, 103739; octubre, 123952 y noviembre, 129042.

Como procedimientos de análisis para este artículo realizamos: a) un primer paso de estadística descriptiva de la producción de noticias y de la cantidad de seguidores que tiene cada una de las cuentas de Twitter sobre 6 meses de emisiones informativas y b) un procedimiento de clasificación de noticias mediado por minería de texto y tratamiento automático del lenguaje sobre la producción informativa del mes de Octubre de 2015. Se decidió hacer un experimento acotado a un mes para probar experimentalmente algoritmos que permitan luego procesar mayores volúmenes de información.

En la primera etapa, los procesos de *Crawling y Scraping* permitieron la extracción de metadatos (fecha y nombres de medios, principalmente) de 736.538 informaciones compiladas. En la segunda etapa, apoyados en la herramienta UIMA (Unstructured Information Management Architecture), se realizaron pretratamientos informatizados propios de la programación de lenguaje natural: a) tokenización, b) análisis gramatical, c) lematización, y d) extracción de palabras claves. En primer lugar, la **tokenización** o proceso de análisis lexicográfico permite discriminar el conjunto de posibles secuencias de caracteres que constituyen un token o lexema. En segundo lugar, el **análisis gramatical** automatizado permite identificar la categoría gramatical de las palabras (nombres, adjetivos, verbos, etc.), por ejemplo: "*Vamos a seguir con el plan de contingencia*", donde "*Vamos*" es el verbo y no debe confundirse con *Vamos* la ciudad en Grecia. Para este caso, el algoritmo de clasificación de secuencias escogido es Hidden Markov Model (HMM), y para el caso del español el modelo se ha entrenado con el corpus "*Spanish* CRATER". En tercer lugar, la **lematización** (lema se refiere al ítem lexical) permite generalizar la forma de las palabras, por ejemplo: *movilizaciones* se puede generalizar como *movilización*; *acusó* se generaliza como *acusar*, etc. En este caso se utiliza un algoritmo basado en reglas específicas para el español. En cuarto lugar, la **extracción de palabras claves** ayuda a identificar las palabras que mejor resumen el texto, utilizando: a) una lista de frecuencias de palabras del español de Chile (Sadowsky 2012), y b) un algoritmo basado en la medida TF-IDF, que relaciona la frecuencia de aparición de un tema en un texto determinado respecto de su frecuencia en una colección de documentos.

Una vez realizado el pre-procesamiento de los datos, se realiza un análisis de las noticias a partir de dos nuevos procesos: a) la clasificación del tema de las noticias mediante un algoritmo de aprendizaje supervisado, y b) la extracción y desambiguación de los nombres de localidades mencionados en la noticia. El proceso de análisis y visualización completo se describe en la *Figura 2.*

### 3.2 Metodología para clasificar el tema

El corpus de entrenamiento utilizado es Wikinews ES, que consiste en 500 noticias etiquetadas en español. Además se considera la representación vectorial de los textos: lista de lemas y su TF-IDF. El algoritmo de aprendizaje supervisado utilizado para la clasificación es Support Vector Machine (SVM) (Joachims 2002) con la herramienta Weka (Waikato Environment for Knowledge Analysis).

**Evaluación del clasificador**: el clasificador se evalúa sobre el corpus de entrenamiento (validación cruzada considerando 10 conjuntos), calculando las medidas de precisión y exhaustividad.

**Precisión (C)**: es la tasa de textos clasificados en la clase C por el clasificador (bien clasificados), versus textos clasificados C por el clasificador. Esta medida permite cuantificar el **ruido** en la clasificación.

**Exhaustividad (C)**: es la tasa de textos clasificados en la clase C por el clasificador (bien clasificados), versus textos clasificados C en realidad. Esta medida permite cuantificar el **silencio** en la clasificación.

*Figura 2.* Esquema de extracción, análisis y visualización de datos textuales

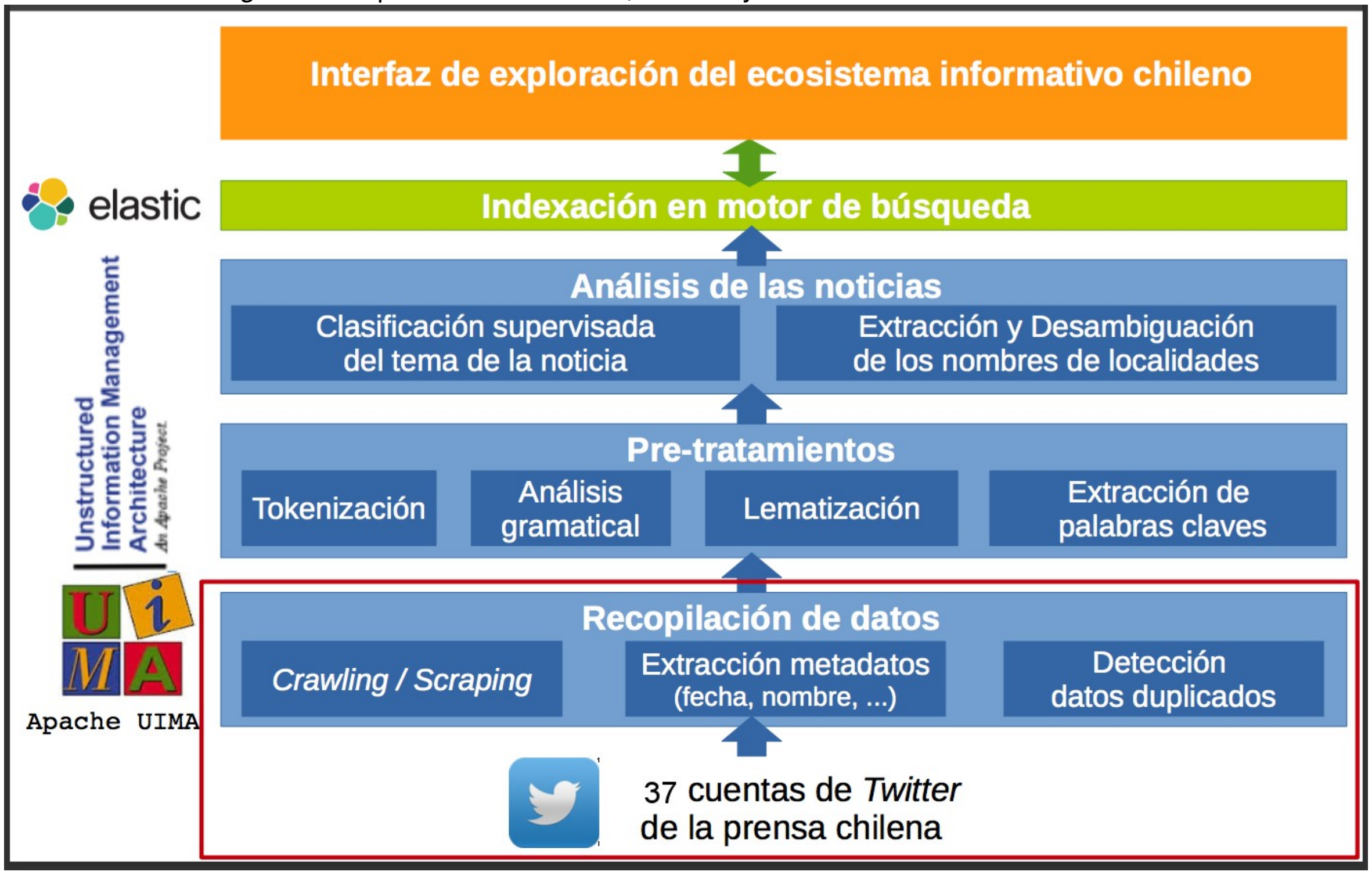

### 3.3 La extracción de localidades

Además de la necesidad de clasificar los temas de referencia de la prensa chilena, resulta interesante conocer si cuando se alude a esos temas se hace referencia a hechos noticiosos ocurridos en una variedad de lugares geográficos o existe una concentración homogénea de referentes. Para el proceso de extracción de los nombres de las localidades se aplicó un segundo nivel de análisis del contenido textual bajo las siguientes reglas metodológicas.

**Contraste frente a recurso *GeoNames.org***: se trata de una base de datos geográfica multilingüe, entre cuyas entradas se encuentran: países, regiones, ciudades, pueblos, lagos, calles, etc., y a su vez presenta coordenadas geográficas de latitud y longitud.

**Indexación de los nombres de localidades en español (MongoDB – 7 Go):** que contiene países (197), ciudades o pueblos (~3.000.000), contrastando en base a un algoritmo de búsqueda de palabras/grupos de palabras en las noticias.

### 3.4 Desambiguación

Evidentemente la identificación de localidades no está exenta de ambigüedades como, por ejemplo: el mismo nombre de la ciudad chilena Valdivia puede tener lugar en otros países del mundo (Colombia, Uruguay, Ecuador, etc.), y, a su vez, muchas palabras en español corresponden a ciudades existentes en el mundo, por ejemplo: la conjugación de tercera persona plural del verbo caer “caen”, es un lexema idéntico a la ciudad francesa Caen. Ante estas ambigüedades conservamos solamente ciudades/pueblos que se encuentran en el país de la noticia.

Finalmente, gracias a Elasticsearch y Kibana, se pudo realizar la indexación de las informaciones compiladas y procesadas. Con ello fue posible ofrecer una interfaz de exploración del ecosistema informativo de medios chilenos en Twitter, lo que facilita la visualización de datos con distintas características.

*Figura 3.* Interfaz de visualización de datos

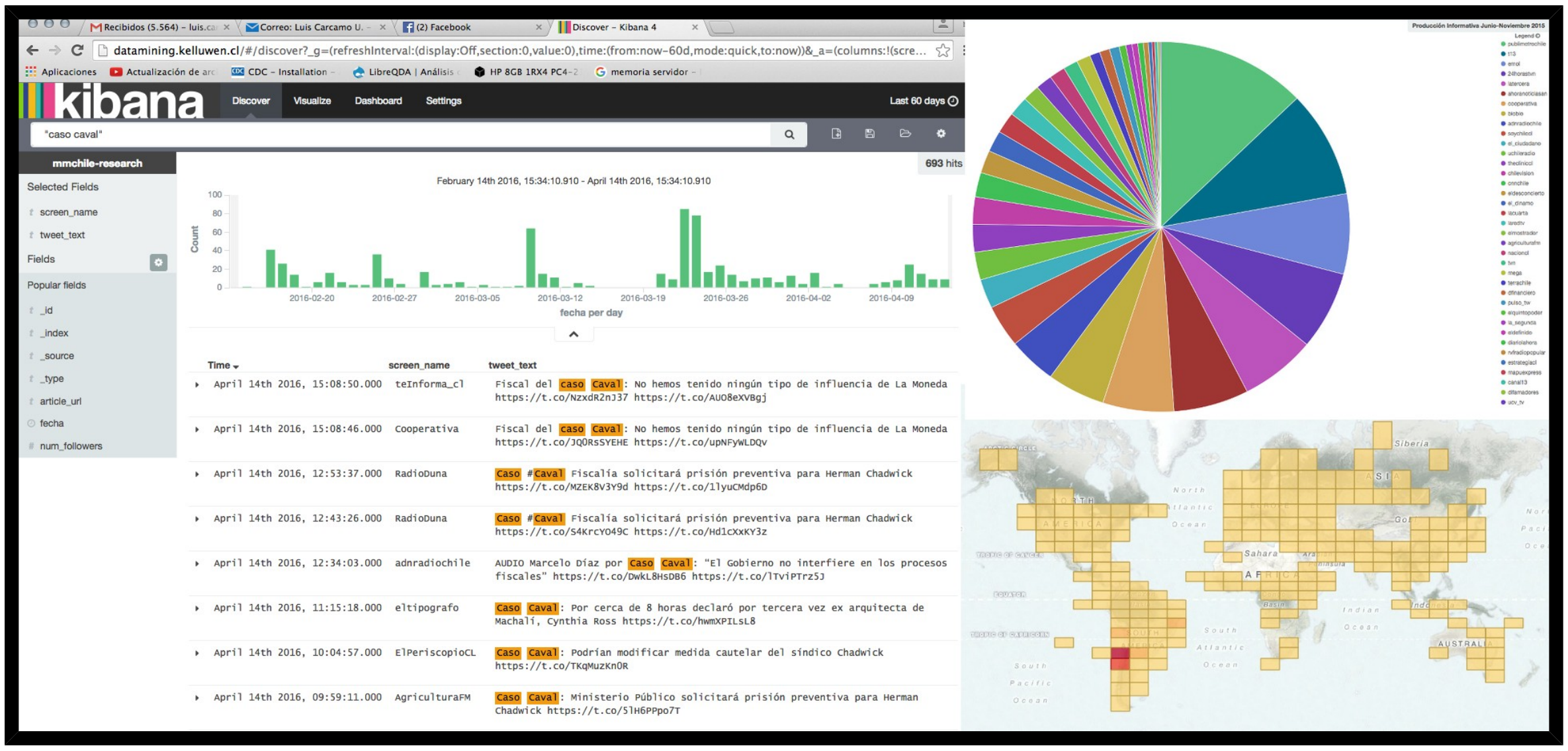

## 4. Resultados

A continuación presentamos los diferentes resultados del análisis realizado. Abordaremos la producción de información en el tiempo, las tendencias temáticas y la cobertura geográfica de los hechos noticiosos.

### 4.1 Volúmenes, ciclos de producción de información y audiencias potenciales

Los datos capturados dan cuenta de una emisión mensual promedio de 120.259,6 noticias al mes. La mayor variación corresponde la mes de septiembre, pues se emitieron 103.739 noticias. Dicha baja se explica básicamente porque septiembre es un mes con una mayor cantidad de días no laborables por los feriados de fiestas patrias y con ello las dinámicas de los turnos de prensa parece disminuir entre los responsables de social media. Al margen de la incidencia del mes de septiembre, en el Gráfico 1 se aprecia cierta estabilidad en la emisión de tweets por parte de los medios de comunicación.

*Gráfico 1.* Emisión global de tweets entre junio y noviembre de 2015

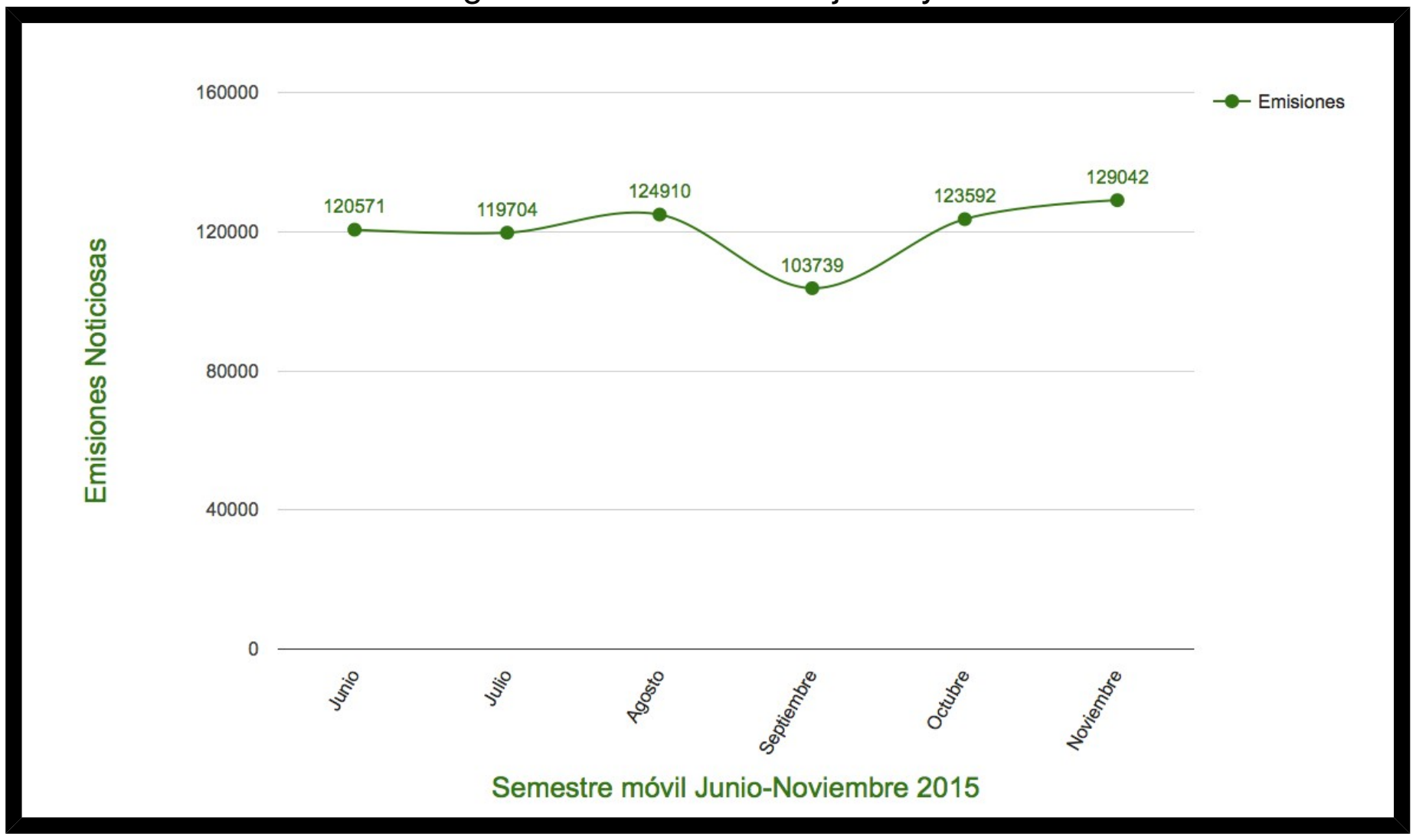

A partir del análisis de los datos específicos del mes de octubre de 2015 se pudo observar que los 37 de medios seguidos por el crawler produjeron 123.952 emisiones en Twitter. De ese total, 65.572 correspondían a noticias diferentes en el mes de octubre de 2015 con un promedio aproximado de ~2.000 por día (Gráfico 2). La diferencia se explica por una tendencia de algunos medios a repetir emisiones con una misma noticia, tweets sin un link o con un link roto que no se pudo minar. La producción de informaciones tuvo un comportamiento bastante estable con incrementos de lunes a viernes y descensos en los fines de semana. Se marcan claramente dos excepciones dadas por los enfrentamientos de la selección chilena de futbol en las eliminatorias del mundial, eventos que se reflejan los días 8 y 13 de octubre.

*Gráfico 2.* Emisión de informaciones en octubre de 2015

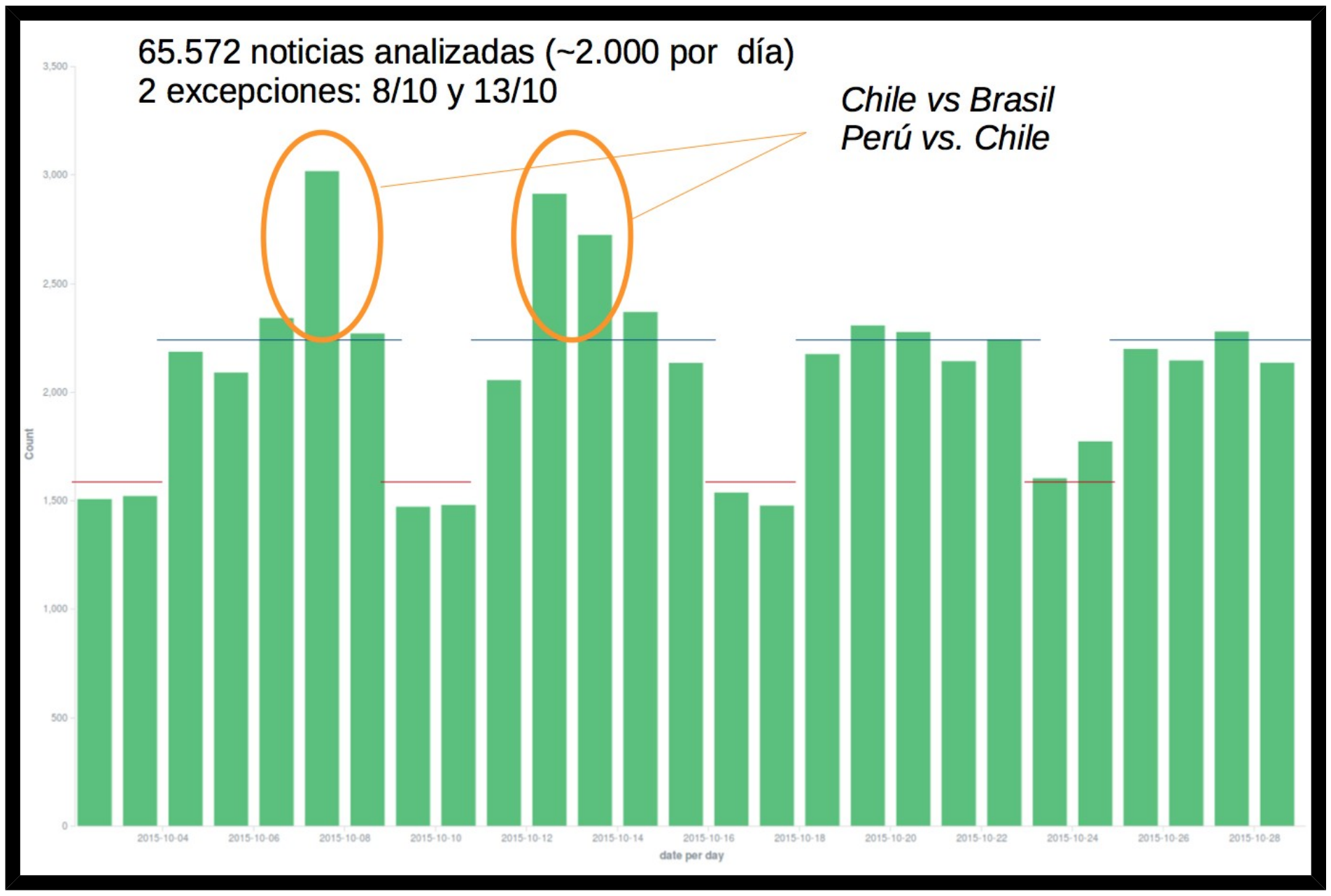

El Gráfico 3 expone que los medios con mayor cantidad de emisiones son medios que tienen producción análoga y digital. Aunque se puede observar una variabilidad en la que alternan medios de prensa en papel y programas televisivos (Tele 13, Canal 24 Horas y Ahora Noticias), radioemisoras (Cooperativa, Bio Bio y ADN Radio Chile) y prensa de papel (Publimetro, Emol, La Tercera y SoyChile). Se puede apreciar cómo se concentra la emisión de informaciones hacia Twitter en los 10 primeros medios. Medios alternativos tales como El Ciudadano, Radio Universidad de Chile, The Clinic y El Desconcierto, El Dínamo y El mostrador ocupan un lugar secundario en términos de producción de información.

*Gráfico 3.* Emisión de tweets por cada medio entre junio y noviembre de 2015

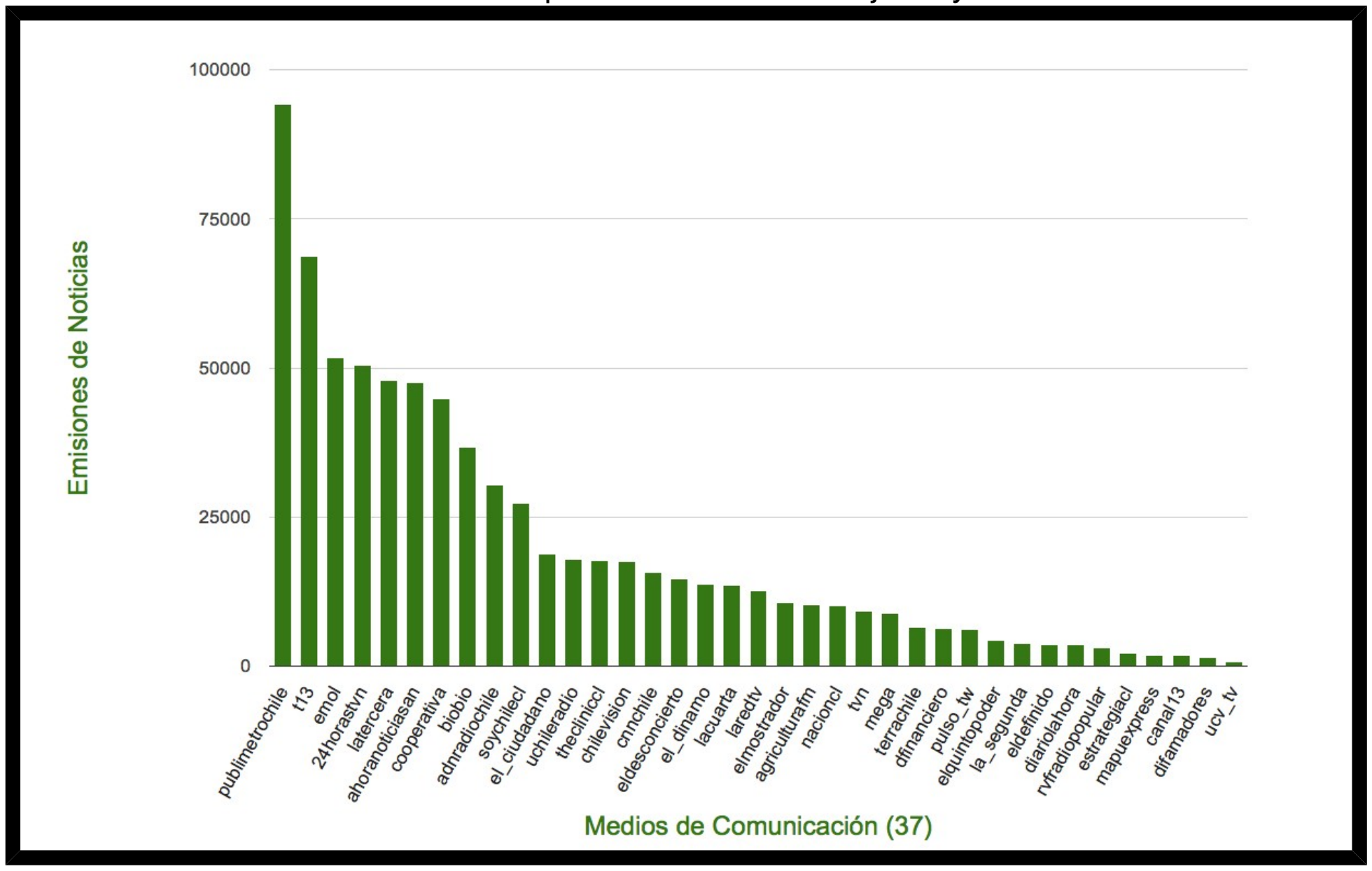

Otra variable importante de revisar dice relación con las potenciales audiencias de los medios de comunicación, que no necesariamente se correlaciona con la producción de información que los medios hacen. El Gráfico 4 permite observar que, de acuerdo al número de seguidores, se reconocen claramente tres grupos de medios. La primera clase de medios supera los 2.000.000 de seguidores y en ella participan medios como Canal 24 Horas, TVN, Tele13, CNN Chile, Bio Bio y Cooperativa. En una segunda clase cohabitan los medios que tienen más de 500.000 seguidores y en ella están La Tercera, Emol, ADN Radio Chile, Chilevisión, La Cuarta, Ahora Noticias, El Mostrador, The Clinic, Terra, Publimetro y Mega. En un tercera clase, con menos de 500.000 seguidores, participan La Segunda, La Red, La Nación, La Hora, El Dínamo, El Ciudadano y Soy Chile, entre otros.

*Gráfico 4.* Clases de medios según número de followers en Twitter

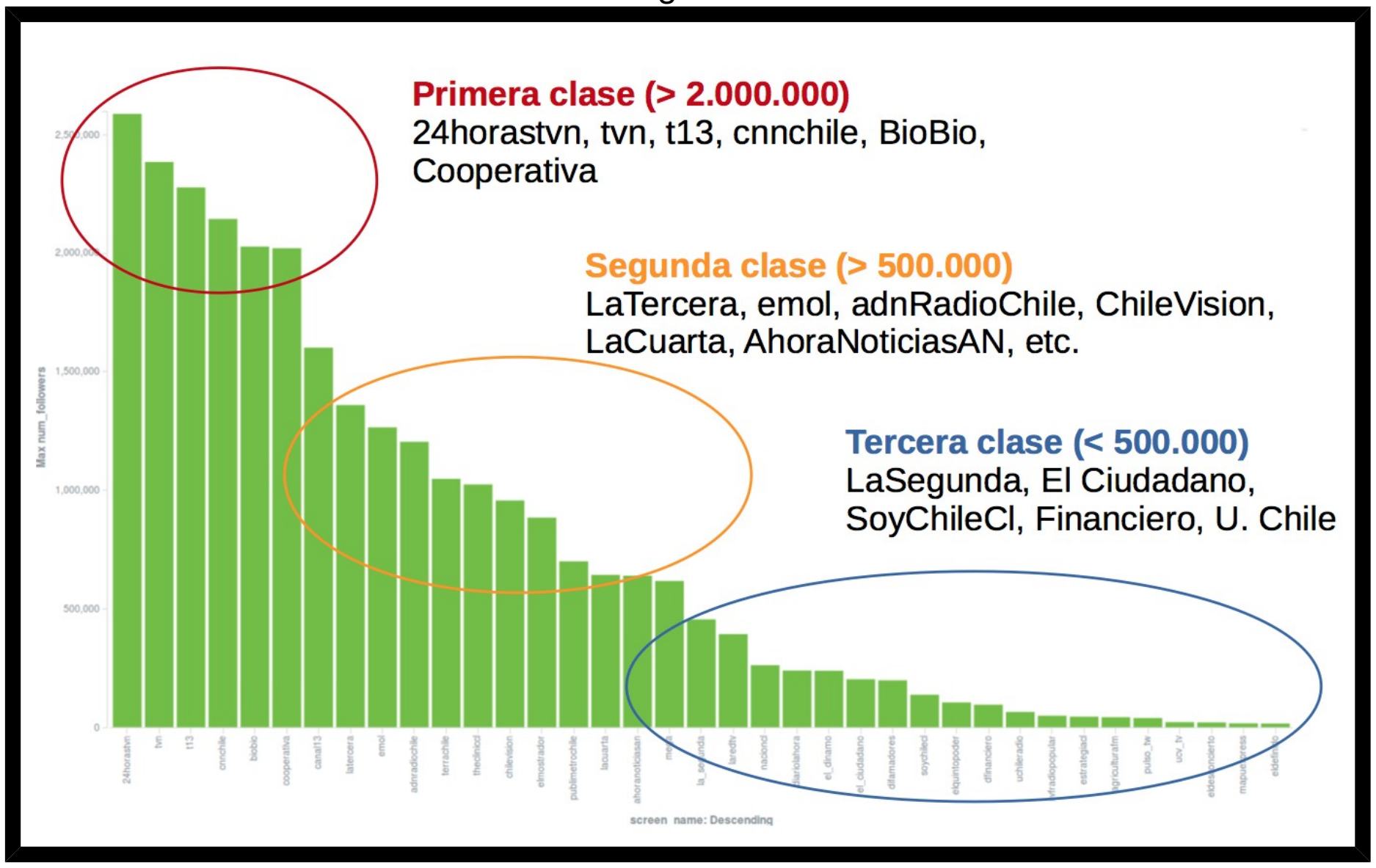

## 4.2 Análisis mediante clasificación de los contenidos de la noticias

En una primera fase, se utilizaron los datos de entrenamiento descritos en la metodología, que permitieron probar un modelo de clasificación de las noticias en los siguientes diez temas posibles: *accidentes, deportes, ecología, economía, entretenimiento, judicial, política, salud, sociedad, tecnología.* Como resultado de este entrenamiento, se obtuvo un modelo de clasificación cuya validez se puede apreciar en la Tabla 1, la cual presenta las medidas de precisión y exhaustividad calculadas. Se puede observar que el modelo se comporta con mayor precisión en los temas de salud, judicial y accidentes, y con mayor exhaustividad en los temas de deporte y salud.

*Tabla 1.* Validación de precisión y exhaustividad de la clasificación de noticias

| **Temas** | **Precisión** | **Exhaustividad** |
|---|---|---|
| Accidentes | 95,8% | 88,5% |
| Deportes | 89,0% | 98,1% |
| Ecología | 71,4%* | 83,3% |
| Economía | 64,4%* | 76,3%* |
| Entretenimiento | 77,8%* | 84,0% |
| Judicial | 97,0% | 88,9% |
| Política | 91,7% | 84,6% |
| Salud | 100% | 96,3% |
| Sociedad | 88,2% | 83,3% |
| Tecnología | 94,7% | 75,0%* |

*Valores cercanos o menores al 75%

Con este modelo se procedió a la clasificación de noticias del mes de octubre de 2015 (65.572 noticias minadas). Con este proceso se pudo observar que más del 50% de las noticias difundidas por los medios de comunicación en Twitter corresponden a Deporte + Entretención (Gráfico 5). Se trata de una situación que se repite al interior de la mayoría de los medios de comunicación seguidos, exceptuándose de esta tendencia medios como: El Desconcierto, Radio Universidad de Chile, El Mostrador, El Financiero, Agricultura, Pulso y Diario Estrategia.

*Gráfico 5*. Distribución de las informaciones en octubre de 2015. Tendencia Deporte + Entretenimiento

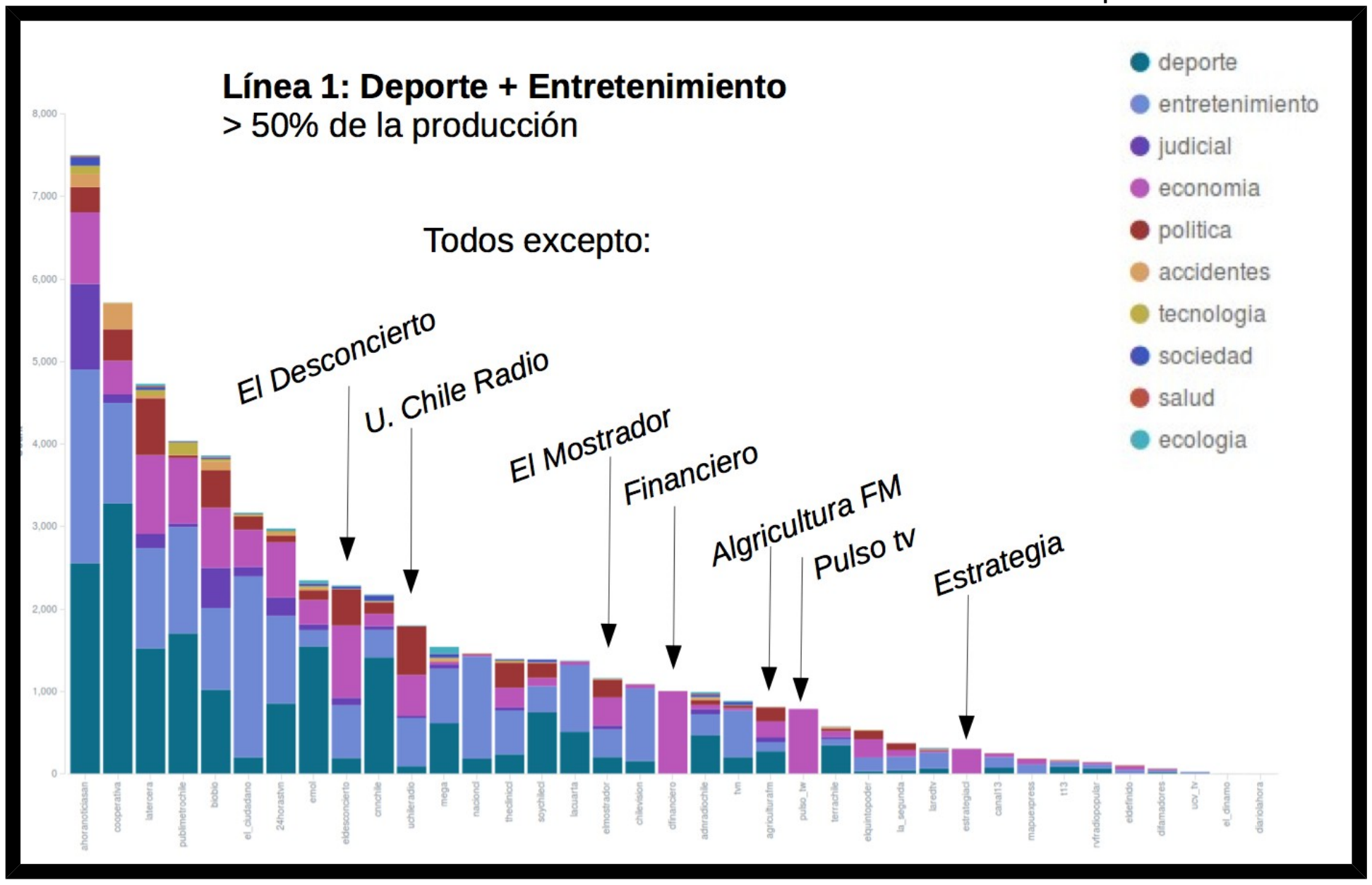

Si bien la tendencia editorial mayoritaria está marcada por el gran número de medios que dedican más del 50% de su producción a deporte y entretenimiento, también se puede observar una segunda tendencia que releva sobre un 25% de su producción a Economía + Política (Gráfico 6). En esta línea se ubican medios tales como La Tercera, Radio Bio Bio, El Desconcierto, Radio Universidad de Chile, The Clinic, El Mostrador, El Financiero, Pulso, El Quinto Poder y Diario Estrategia.

*Gráfico 6.* Distribución de las informaciones en octubre de 2015. Tendencia Economía + Política

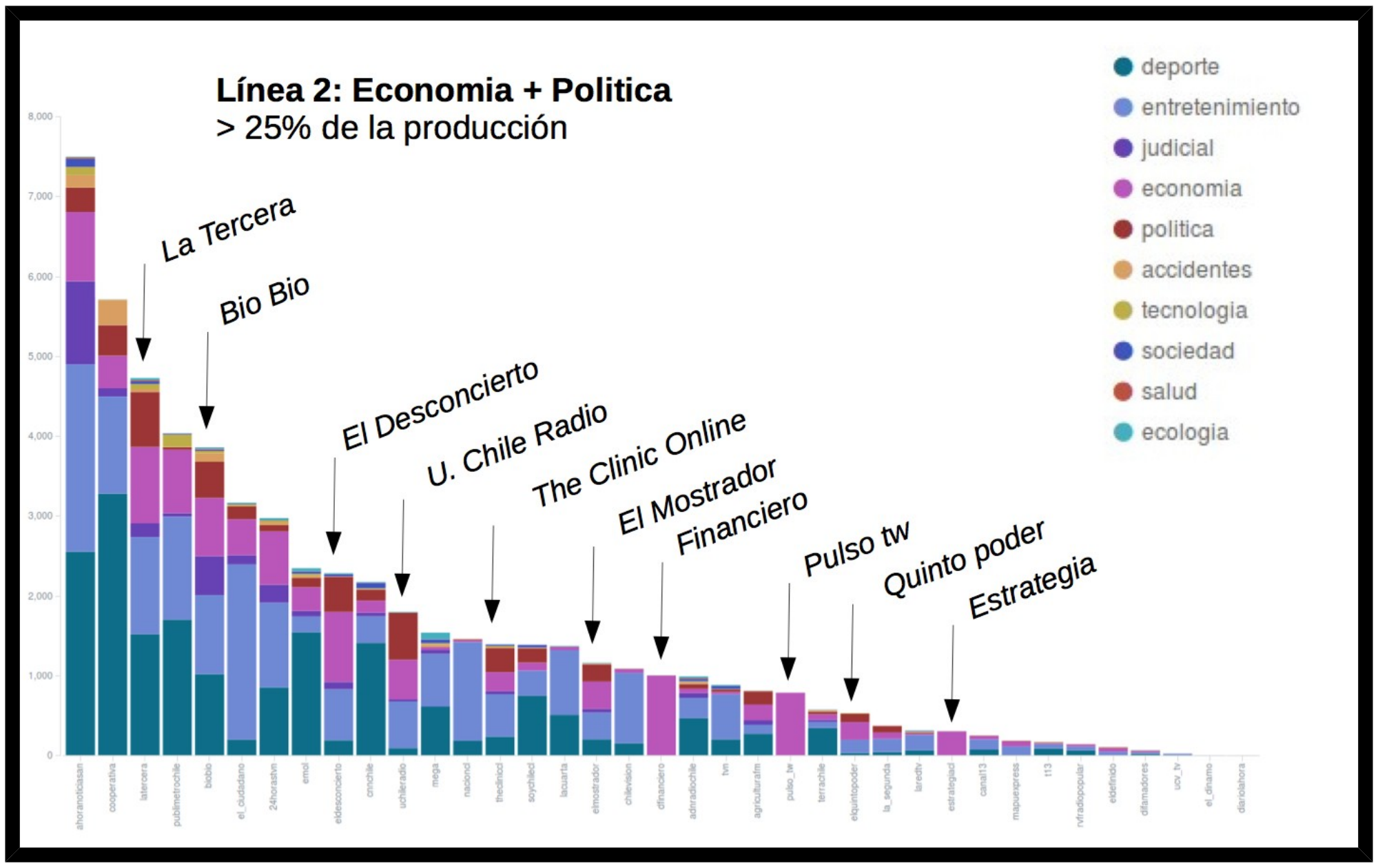

Se presenta también una tercera tendencia de medios que dedican más de un 10% de su producción informativa a eventos judiciales (homicidios, tráfico de drogas, asaltos, entre otros). En este grupo se ubican Ahora Noticias, Radio Bio Bio y ADN Radio Chile (Gráfico 7).

*Gráfico 7.* Distribución de las informaciones en octubre de 2015. Tendencia Judicial

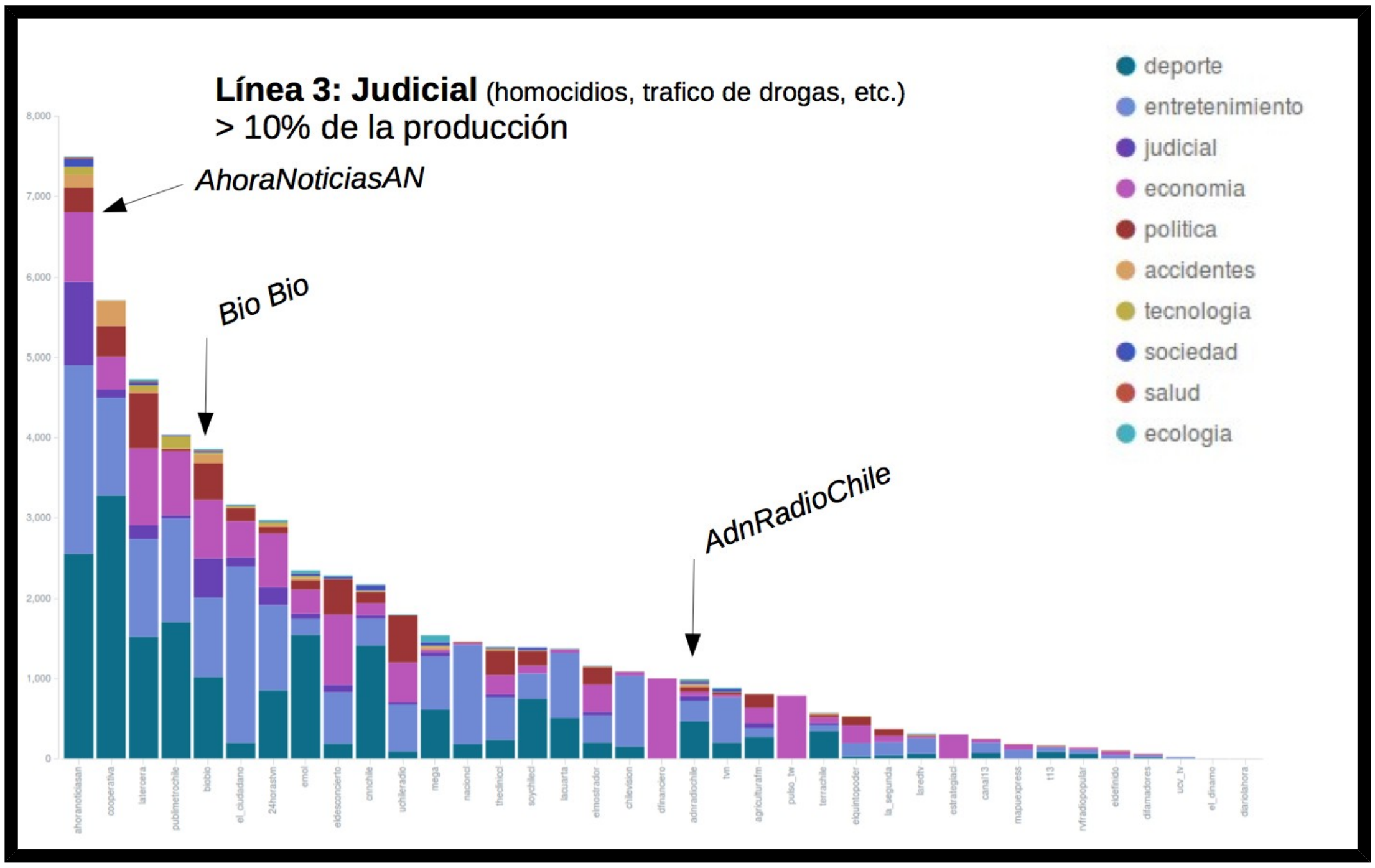

## 4.3 Cobertura geográfica de la información

Los datos del mes de octubre de 2015 dieron cuenta de una esperable centralización de las noticias que refieren a Santiago de Chile, pero también se expresa cómo un fenómeno natural (temblor en La Serena) es capaz de llamar la atención de las pautas informativas, o cómo eventos de deportes y entretención son capaces de llamar fuertemente la atención de los medios chilenos aun cuando sucedan en otro país, por ejemplo, Argentina (Mapa 1). Así también, queda en evidencia la ausencia de informaciones que refieran a las regiones de Aysén y Magallanes.

*Mapa 1.* Distribución geográfica de las informaciones en octubre de 2015

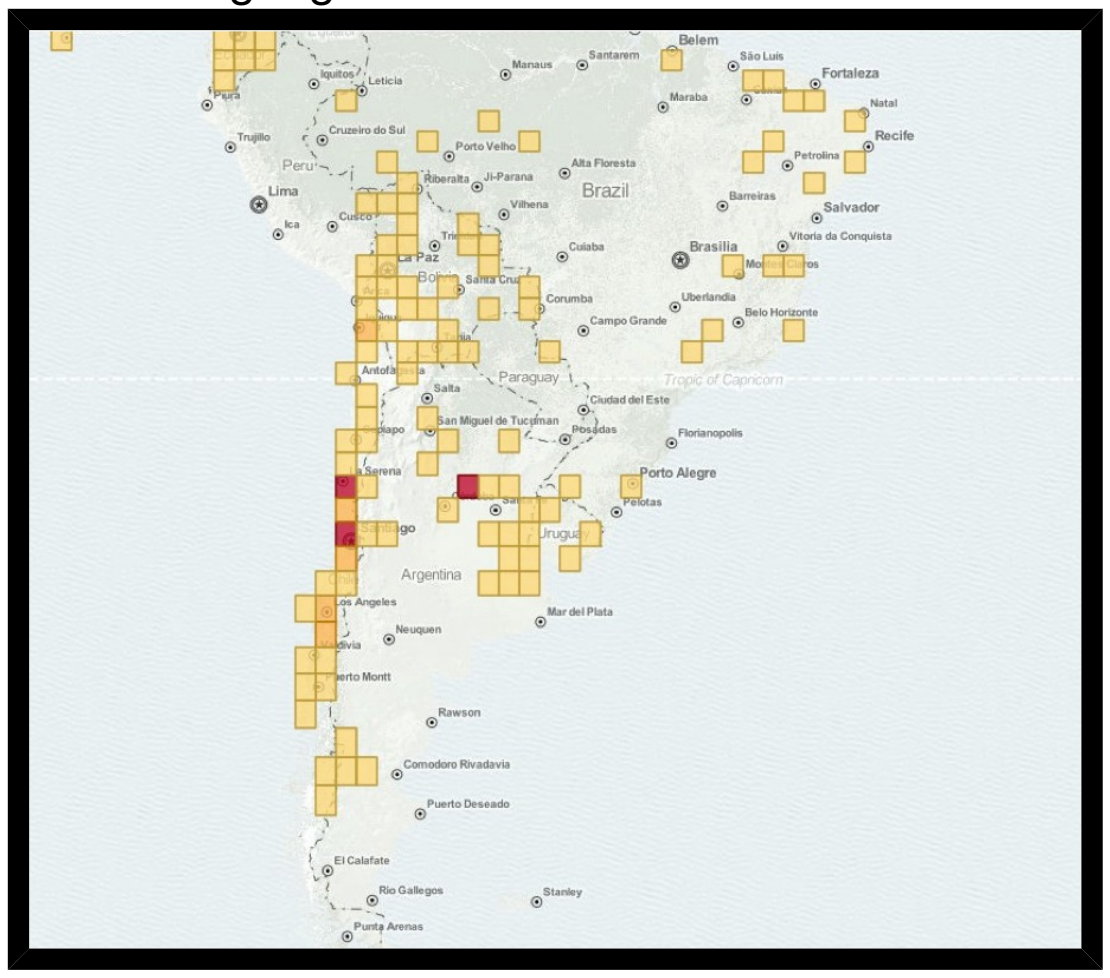

A nivel global se pudo observar que los medios chilenos incluyen en sus pautas informativas noticias de gran parte del mundo (Mapa 2). Se constata, como era esperable, un fuerte componente de noticias nacionales. La presencia de noticias referentes a Perú y Brasil se explica por las eliminatorias al mundial (fenómeno también reflejado en el Gráfico 2). También se expresa una considerable cantidad de noticias referidas a Argentina (tendencia que se explica por una conjunción de noticias deportivas, políticas y de farándula de dicho país).

*Mapa 2.* Distribución geográfica de las informaciones en octubre de 2015

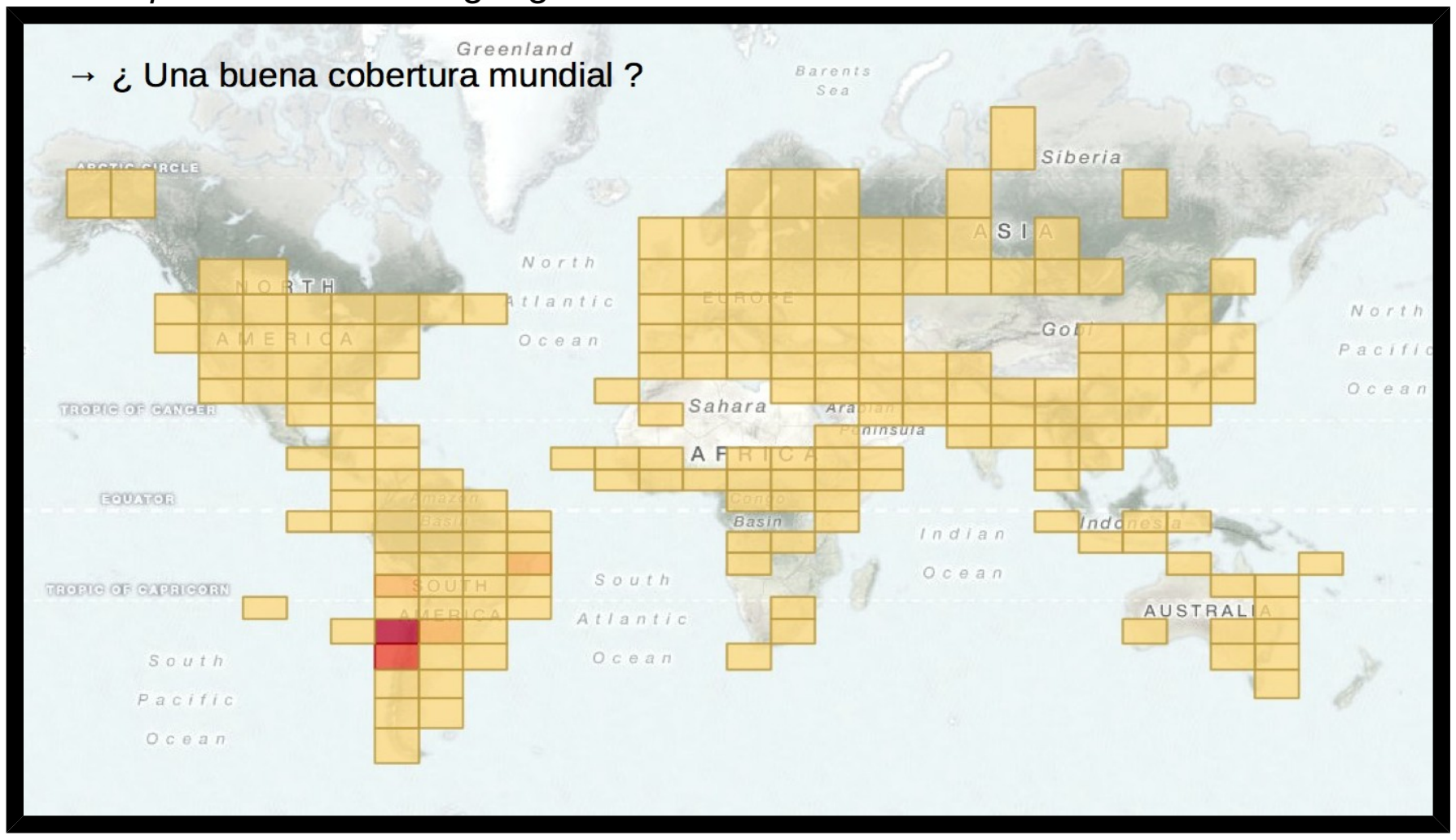

Obviamente la configuración de la cobertura global se constituye de las coberturas parciales de los distintos medios chilenos. Así, por ejemplo, sobre temas de política, La Tercera ostenta una cobertura global mayor que Radio Cooperativa, tal como se observa en los Mapas 3 y 4.

*Mapas 3 y 4.* Distribución geográfica de las informaciones según las localidades del mundo mencionadas por Cooperativa y LaTercera

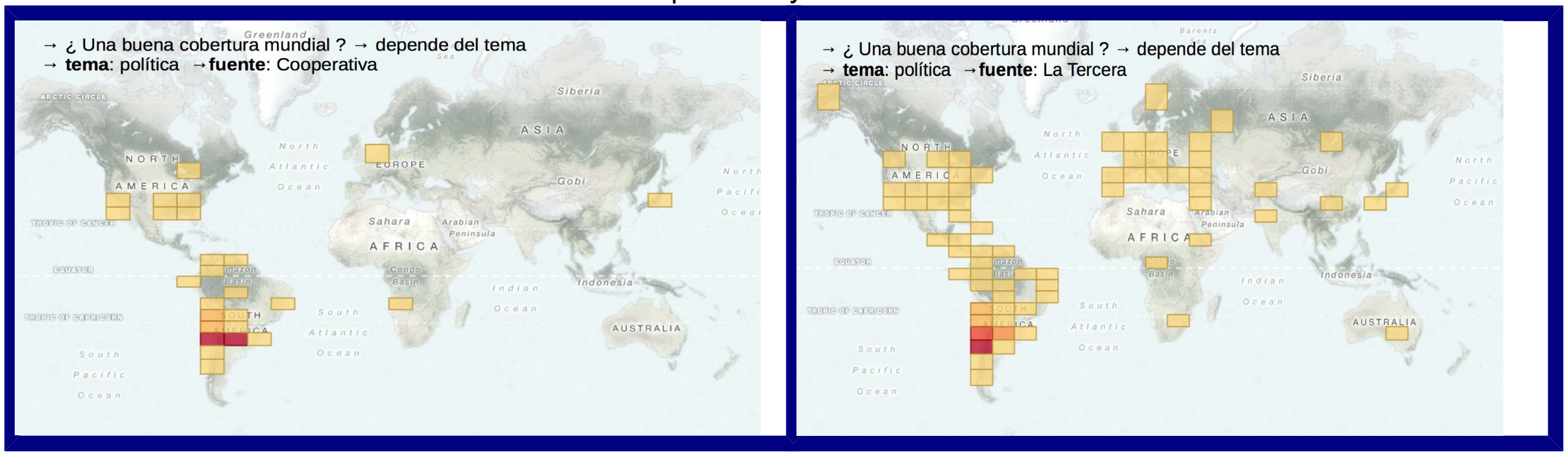

## 5. Conclusiones

Siendo un primer estudio en el contexto de una investigación con un plazo de cuatro años, ya es posible hacer una serie de observaciones generales:

- Existe una gran disparidad en la emisión de informaciones que hacen los medios de comunicación del ecosistema informativo chileno: 10 medios producen el 68% de las emisiones y 27 generan el 32% de los tweets en el período junio-noviembre de 2015.
- La producción de la información no se correlaciona directamente con las potenciales audiencias alcanzadas. Por ejemplo, un medio como Publimetro que emite muchos contenidos diariamente no se instala como uno de los medios con mayor cantidad de seguidores en Twitter.
- Hasta ahora podemos distinguir tres clases de medios. En primer lugar, se encuentra un pequeño grupo de **audiencia potencial alta** constituido por canales de televisión informativos y radioemisoras con audiencias mayores a 2.000.000 de seguidores. Un segundo grupo se caracteriza por una **audiencia potencial media** constituido por periódicos tradicionales con audiencias mayores a 500.000 seguidores. Finalmente, está el grupo de **audiencia potencial baja** en el que cohabitan periódicos con foco temático específico y medios de pauta informativa alternativa con menos de 500.000 seguidores.
- También se pudo apreciar que la cobertura geográfica de las informaciones resulta bastante completa al explorar el conjunto de medios seguidos (37), pero se pudo observar bastante variación entre la cobertura que hace cada medio. Además queda en evidencia que existen territorios como Magallanes sobre los que las noticias chilenas no hacen mayores referencias.
- Las tendencias temáticas informativas parecen fuertemente orientadas a deportes y entretenimiento (más de un 50% en 30 medios), y marginalmente orientadas a sucesos policiales y judiciales (sólo en 3 medios superan el 10% de su producción informativa). Esta última tendencia parece marcar un cambio con respecto a la tradición chilena informativa que siempre ha destinado gran cobertura a los sucesos policiales, sin embargo, este cambio puede deberse a una tendencia editorial sólo aplicada a la información del público destinatario de Twitter, que etaria y socialmente está definido en un 65% como un público joven-adulto de entre 15 y 30 años aproximadamente.

- La centralización de la cobertura informativa parece mantenerse como una constante de la prensa chilena. En el ámbito nacional existen territorios de los que los medios no hablan (Magallanes y Aysén), y a nivel mundial el conjunto de los medios parece ofrecer una buena cobertura, pero al analizar los contenidos de cada medio se observan grandes variaciones.

## Bibliografía

Azocar, A.; Scherman, A.; Arriagada, A.; Pardo, J. & Becerra, A. 2010. Primer Estudio Nacional sobre Lectoría de Medios Escritos. Universidad Diego Portales. En: http://www.udp.cl/investigacion/repo_detalle.asp?id=76 (Consultado en Febrero de 2016)

Arriagada, A; Correa, T.; Scherman, A. & Abarzúa, J. 2014. Diarios de Vida de las Audiencias Chilenas. Universidad Diego Portales y Conicyt Pluralismo. En: http://www.diariosdevida.udp.cl//wp-content/uploads/2015/05/Informe_Final_Diarios_de_Vida_de_las_Audiencias.pdf (Consultado en Febrero de 2016)

Baeza-Yates, R. 2009. "Tendencias en minería de datos de la Web". El profesional de la información, 18(1): 5-10.

Baeza-Yates, R y Pasi, G. 2011. "Special issue of The Journal of Information Retrieval on web mining for search". Information Retrieval, 14(3): 213-214.

Baeza-Yates, R. & Sáez-Trumper, D. 2015. "Wisdom of the Crowd or Wisdom of a Few? An Analysis of Users' Content Generation". Proceedings of the 26th ACM Conference on Hypertext & Social Media. Pp. 69-74

Cabalin, C. (2014). "Estudiantes conectados y movilizados: El uso de Facebook en las protestas estudiantiles en Chile". Comunicar, 43: 25-33.

Cárcamo-Ulloa, L., & Saez-Trumper, D. 2013. "¿ Cambian las hegemonías periodísticas en las redes sociales? Prensa chilena en Facebook". Revista Nhengatu 1(1): s/p. En: http://www.nhengatu.org/revista/index.php?journal=nhengatu&page=article&op=view&path[]=6 (Consultado en Febrero de 2016)

Cárcamo-Ulloa, L. & Sáez-Trumper, D. 2014. "Medios de Comunicación de Masas en Facebook. Comparativa de Chile y España". XX Congreso Internacional de la Sociedad Española de Periodística (SEP) 2014. Barcelona, España.

Cardoso, G. 2014. "Movilización social y medios sociales". Vanguardia dossier 50: 16-23.

ComScore. 2013. Futuro Digital Chile 2013. En: https://www.comscore.com/lat/Insights/Presentations_and_Whitepapers/2013/2013_Chile_Digital_Future_in_Focus. (Consultado en Febrero de 2016).

Corrales. O. & Sandoval, J. 2005. "Medios de comunicación, pluralismo y libertad de expresión", Colección ideas, Cuadernos de trabajo Fundación Chile XXI. Nº 53.

CNTV. 2014. Movilizaciones Estudiantiles. Percepciones de los Jóvenes. Dpto de Estudios, Consejo Nacional de Televisión. http://www.cntv.cl/movilizaciones-estudiantiles-percepcion-jovenes/prontus_cntv/2014-04-04/154750.html (Consultado en Febrero de 2016)

Díaz-Nosty, B. 2013. La prensa en el nuevo ecosistema informativo "¡Que paren las rotativas!. Fundación Telefónica. Barcelona: Ariel.

García, R., Botella, F., & Marcos, M. C. 2010. "Hacia la arquitectura de la información 3.0: pasado, presente y futuro". El profesional de la información 19(4): 339-347.

Guzman, J. & Poblete, B. 2013. On-line relevant anomaly detection in the twitter stream: an efficient bursty keyword detection model. In*Proceedings of the ACM SIGKDD Workshop on Outlier Detection and Description* (pp. 31-39).

Halpern, D. (2014). Social TV en Chile: Hábitos y tendencias. VTR - Facultad Comunicaciones UC – TrenDigital. En: http://www.iab.cl/social-chile-habitos-tendencias-facultad-comunicaciones-tren-digital-abril-2014/ (Consultado en Febrero de 2016)

IAB. 2012. Uso de redes sociales en Chile. Interactive Advertising Bureau (IAB). En: http://www.iab.cl/uso-de-redes-sociales-en-chile-octubre-2012/ (Consultado en Febrero de 2016).

Joachims, T. 2002. Learning to classify text using support vector machines: Methods, theory and algorithms. Kluwer Academic Publishers.

Kurth, M. 1993. "The limits and limitations of transaction log analysis", Library Hi Tech 11(2): 98-104.

Marcos, M. C., & González-Caro, C. 2010. "Comportamiento de los usuarios en la página de resultados de los buscadores. Un estudio basado en eye tracking". El profesional de la información 19(4): 348-358.

Masip, P., Díaz-Noci, J., Domingo, D., Micó-Sanz, J. L., & Salaverría, R. 2010. "Investigación internacional sobre ciberperiodismo: hipertexto, interactividad, multimedia y convergencia". El profesional de la información 19(6): 568-576.

Morales, C., Soto, S. & Martínez, C. 2005. “Estado actual de la aplicación de la minería de datos a los sistemas de enseñanza basada en web”. Actas del III Taller Nacional de Minería de Datos y Aprendizaje, TAMIDA2005, 49-56.

Newman, N. 2013. Reuters Institute Digital News Report 2013: Tracking the Future of News. En: https://reutersinstitute.politics.ox.ac.uk/sites/default/files/Digital%20News%20Report%202013.pdf (Consultado en Febrero de 2016)

Pardo, N. 2012a. “Metáfora multimodal: Representación mediática del despojo”. Forma y Función, 25(2): 39-61.

Pardo, N. 2012b. Discurso en la Web: Pobreza en Youtube. Universidad Nacional de Colombia. Editorial Grafiweb. Bogotá Colombia.

Peters, T. 1993. “The history and development of transaction log analysis”. Library hi tech, 42: 41–66.

Puente, S. & Grassau, D. 2009. “Informaciones regionales: calidad y presencia en la agenda noticiosa nacional”. Cuadernos de Información 25: 29-38.

Ramirez, J.D. 2009. “La concentración de la propiedad radial en Chile: las exigencias de nuevos paradigmas entre Globalidad y Localidad”. Revista Redes.com 5: 309-327.

Sadowsky, S. & y Martínez, R. 2012. *Lista de Frecuencias de Palabras del Castellano de Chile (Lifcach)*. Versión 2.0. Base de datos electrónica. http://sadowsky.cl/lifcach.html

Sánchez, H. & Méndez, S. 2013. “¿Perfiles profesionales 2.0? Una aproximación a la correlación entre la demanda laboral y la formación universitaria”. Estudios sobre el mensaje periodístico 19: 981-993.

Stambuk, P. 1999. “El desequilibrio informativo en Chile: discriminación de la noticia regional en la prensa nacional”. Revista Latina de Comunicación Social 14 : s/p. En http://www.ull.es/publicaciones/latina/a1999c/133valpara.htm (Consultado en Febrero de 2016)

Saez-Trumper, D., Castillo, C., & Lalmas, M. 2013. “Social media news communities: gatekeeping, coverage, and statement bias”. Proceedings of the 22nd ACM international conference on Conference on information & knowledge management pp. 1679-1684.

Sáez-Trumper, D. 2011. “La información en Internet: Breve estado del arte para discutir el poder de los usuarios v/s los medios tradicionales de comunicación en la red”. Revista Austral Ciencias Sociales 20: 71-79.

Tejedor-Calvo, S. 2010. “Web 2.0 en los ciberdiarios de América Latina, España y Portugal”. El profesional de la información 19(6): 610-619.

Yang, J., & Leskovec, J. 2011. “Patterns of temporal variation in online media”. Proceedings of the fourth ACM international conference on Web search and data mining, pp. 177-186.

Vera, S. 2005. Concentración de la propiedad de los medios de comunicación en Chile y sus efectos en el pluralismo informativo. Memoria para optar al grado de Licenciado en Ciencias Jurídicas y Sociales. Universidad Austral de Chile

Vernier, M., Monceaux, L. & Daille, B. 2012. Détection de la subjectivité et catégorisation de textes subjectifs par une approche mixte symbolique et statistique. In Expérimentations et évaluations en fouille de textes : un panorama des campagnes DEFT. Chapitre 7. Collection "Systèmes d'Information et Organisations Documentaires". Grouin C, Forest D (eds) sous la direction de S. Chaudiron. Hermes-Lavoisier.

Vergara, E., Garrido, C., Santibáñez, A., & Vera, P. 2012. “Inversión publicitaria y pluralismo informativo: Una aproximación comparada al caso de la prensa en Chile”. *Comunicación y Medios* 25: 57-70.